\documentclass[
]{ceurart}

\usepackage{listings}
\usepackage{graphicx}
\usepackage{latexsym}
\usepackage[small]{caption}
\usepackage{subcaption}
\usepackage{graphicx}
\usepackage{amsmath,amssymb}
\usepackage{amsthm}
\usepackage{booktabs}
\usepackage{algorithm}
\usepackage{algorithmic}
\usepackage[scr=rsfs]{mathalpha}
\usepackage{adjustbox}
\usepackage{xcolor}
\usepackage{float}
\begin{document}

\copyrightyear{2023}
\copyrightclause{Copyright for this paper by its authors. Use permitted under Creative Commons License Attribution 4.0 International (CC BY 4.0).}

\conference{Aequitas 2023: Workshop on Fairness and Bias in AI $\vert$ co-located with ECAI 2023, Kraków, Poland}

\title{EXTRACT: Explainable Transparent Control of Bias in Embeddings}

\author[1]{Zhijin Guo}[%
email=zhijin.guo@bristol.ac.uk,]
\address[1]{University of Bristol, Beacon House, Queens Road, Bristol, BS8 1QU, UK.}
\author[1]{Zhaozhen Xu}[%
email=zhaozhen.xu@bristol.ac.uk,
]

\author[1]{Martha Lewis}[%
email=martha.lewis@bristol.ac.uk,
url=https://research-information.bris.ac.uk/en/persons/martha-lewis,
]

\author[2]{Nello Cristianini}[%
email=nc993@bath.ac.uk,
url=https://researchportal.bath.ac.uk/en/persons/nello-cristianini,
]
\address[2]{University of Bath, Claverton Down, Bath BA2 7AY, UK }
\begin{abstract}
Knowledge Graphs are a widely used method to represent relations between entities in various AI applications, and Graph Embedding has rapidly become a standard technique to represent Knowledge Graphs in such a way as to facilitate inferences and decisions. As this representation is obtained from behavioural data, and is not in a form readable by humans, there is a concern that it might incorporate unintended information that could lead to biases. We propose EXTRACT: a suite of \textbf{Ex}plainable and \textbf{Tra}nsparent methods to \textbf{C}on\textbf{T}rol bias in knowledge graph embeddings, so as to assess and decrease the implicit presence of protected information. Our method uses Canonical Correlation Analysis (CCA) to investigate the presence, extent and origins of information leaks during training, then decomposes embeddings into a sum of their private attributes by solving a linear system. Our experiments, performed on the MovieLens-1M dataset, show that a range of personal attributes can be inferred from a user’s viewing behaviour and preferences, including gender, age and occupation. Further experiments, performed on the KG20C citation dataset, show that the information about the conference in which a paper was published can be inferred from the citation network of that article. 
We propose four transparent methods to maintain the capability of the embedding to make the intended predictions without retaining unwanted information. A trade-off between these two goals is observed.
\end{abstract}


\maketitle

\thispagestyle{plain}
\pagestyle{plain}

\section{Introduction}
Knowledge graphs are structures that encode entities and the relationships between them, and are useful representations of real world phenomena. Knowledge graphs can be employed in an array of applications, including linguistic representation learning \cite{logan2019barack}, question answering \cite{dai2016cfo}, multihop reasoning \cite{bauer2018commonsense}, or recommender systems \cite{ji2021survey}. 

This study emphasizes \textit{ link prediction }, an inherent task in Knowledge Graph Embedding, that can simplify crucial issues like recommending user actions or answering entity-specific queries. Using user gender as an exemplar of private information, we scrutinize its influence on representations and its ethical implications within AI systems.

Recent studies showed that personal information can be inferred from user behaviour. A study of Facebook users showed that certain private traits (including gender and race) could be reconstructed on the basis of their public ``likes'' \cite{kosinski2013private}; and a study of word embeddings showed that these representations contained information related to gender and race, that might potentially affect the decisions of algorithms that make use of those representations \cite{caliskan2017semantics}.
We investigate this problem with specific application to detecting and removing bias in knowledge graphs.

We address issues, similar to prior word embedding research, concerning vector representations derived from word co-occurrence \cite{pennington2014glove}, recasting this process as a particular facet of the graph embedding problem. The notable ``compositionality" feature in word embeddings signified the potential for analogical reasoning and the risk of undesirable encodings, such as gender biases \cite{mikolov2013distributed}.
It was observed that these distributions also contain additional information including associations and biases that reflect customs and practices. For example, the embeddings of color names were not gender neutral \cite{jonauskaite2021english}, nor were those of job titles or academic disciplines. Engineering disciplines and leadership jobs tended to be represented in a ``more male" way than artistic disciplines or service jobs \cite{jonauskaite2021english,caliskan2017semantics}.

\paragraph{Related Work}
The presence of gender information in word embeddings was already reported in \citet{bolukbasi2016man}, in an article aptly entitled ``Man is to Computer Programmer as Woman is to Homemaker?'', as well as in \citet{mikolov2013distributed}. An interesting possibility is the presence of similar biases in Knowledge Graph embedding, which would lead both to opportunities and challenges, and which would require attention. Our research significantly extends the field by introducing detection methods under the EXTRACT framework. In a ``fair classification" context, \citet{madras2018learning} aimed to predict a label $y$ from data $x$, while ensuring fairness regarding a binary sensitive attribute $a$. Our detection methods are tailored to identify specific dimensions in user embeddings that are associated with private or sensitive attributes. Notably, our linear decomposition system reveals that user behaviour can be approximated as a sum of demographic vectors, uncovering another layer of embedded bias.

When it comes to bias removal, a range of strategies exists. Regularization methods focusing on fairness have been explored \cite{chen2023bias}. The LFR framework \cite{zemel2013learning} aimed to encode non-sensitive attributes while minimizing sensitive ones. Its extension, ALFR, has been adopted in recommendation systems \cite{edwards2015censoring}. Works like \citet{zhu2020measuring} employ adversarial learning for equal score distribution, \citet{bose2019compositional} and \citet{wu2021learning} extended ALFR's scope by imposing compositional fairness. Notably, \citet{fisher-etal-2020-debiasing} used adversarial loss for model neutrality. Our contributions in debiasing offer transparency and interpretability. We employ linear transformations to identify bias direction and leverage first-moment loss to minimize distributional disparities during training.
\paragraph{Our Contributions}

We introduce EXTRACT: Explainable and Transparent ConTrol of bias in embeddings, a two-step process to detect and mitigate bias in Knowledge Graph Embeddings, implemented on the MovieLens 1M and KG20C citation datasets. Our detection methods include logistic classifiers (detect-LC), canonical correlation analysis (detect-CCA), and a linear decomposition (detect-LD). Bias removal leverages linear projection methods (remove-LP and remove-LP-multi) and two retraining methods, remove-FM and remove-FM-multi.

The proposed methods are effective and interpretable, with our novel application of detect-CCA and detect-LD. Specifically, detect-CCA constitutes a novel use of CCA, marking its debut in bias detection. Additionally, detect-LD contributes a unique decomposition of node embedding space into interpretable dimensions, signifying attribute presence or absence - a method not previously implemented. Our methods, remove-LP and remove-LP-multi, can be directly applied to pre-existing embeddings, saving computational resources by eliminating the need for retraining from scratch. Despite slight performance decreases, our methods parallel state-of-the-art results, maintaining conciseness, interpretability, and explainability.

\section{Mathematical Preliminaries}
\label{sec:preliminaries}
\paragraph{Knowledge Graph-Based Recommendation Algorithms}

In a \textit{knowledge graph} $G= (V, \mathcal{F}, A)$, vertices $V$ signify real-world entities, each endowed with a set $A$ of attributes. Relationships between these vertices are captured in a set $\mathcal{F}$ consisting of \emph{facts}. Each fact $f \in \mathcal{F}$ is articulated as a triple $(h,r,t)$, where $r$ belongs to a set of relations $\mathcal{R}$  that serve as direct edges from head $h$ to tail $t$. For instance, considering a set of users and movies as vertices, user ratings form the edges. 

A knowledge graph embedding translates vertices into vectors maintaining graph topology, represented as $(\mathbf{h},\mathbf{R},\mathbf{t}) \in \mathbb{R}^n$. The score function $S (\mathbf{h}, \mathbf{R}, \mathbf{t})$ governs the linking between nodes.
\paragraph{Rating Prediction}

In alignment with \citet{berg2017graph}, we establish a function $P$ that, given a triple of embeddings $(\mathbf{h},\mathbf{R},\mathbf{t})$, calculates the probability of the relation against all potential alternatives.
\begin{eqnarray}
P\left(\mathbf{h},\mathbf{R},\mathbf{t}\right)=\text{SoftArgmax}(S(f)) =\frac{e^{S(f)}}{e^{S(f)}+\sum_{r' \neq r\in \mathscr{R}} e^{S(f')}}
\label{eqn: loss}
\end{eqnarray}
In the above formula, $f =(h,r,t)$ denotes a true triple, and $f'=(h,r',t)$ denotes a corrupted triple, that is a randomly generated one. We use as a proxy for a negative example (a pair of nodes that are not connected). 

Assigning numerical values to relations $r$, the predicted relation is then just the expected value $
\text{prediction} = \sum_{r \in \mathscr{R}} r P\left(\mathbf{h},\mathbf{R},\mathbf{t}\right)$
In our application of viewers and movies, the set of relations $\mathscr{R}$ could be the possible ratings that a user can give a movie. The predicted rating is then the expected value of the ratings, given the probability distribution produced by the scoring function. $S(f)$ refers to the scoring function in \citet{yang2014embedding}.

To learn a graph embedding, we follow the setting of \citet{bose2019compositional} as follows, 

\begin{equation}
        L  = - \sum_{f \in \mathscr{F}} \log \frac{e^{S(f)}}{e^{S(f)}+\sum_{f' \in \mathscr{F}'} e^{S(f')}}
     \label{eq:actual_loss}
\end{equation}
This loss function maximises the probabilities of true triples $(f)$ and minimises the probability of triples with corrupted triples $(f')$.
\paragraph{Entity prediction}
Unlike predicting movie ratings which is the relation in MovieLens 1M, we use a margin based loss \cite{han-etal-2018-openke} to predict the tail entities (which could be paper, affilation and domain) as follows. In this loss function,  $f$ represents for a true triple while $f'$ represents for a corrupted triple, 
\begin{equation}
L=\sum_{f \in \mathcal{F}} \sum_{f^{\prime} \in \mathcal{F}^{\prime}}\left[\gamma+S\left(f^{\prime}\right)-S(f)\right]_{+}
\label{eq: actual_loss_margin}
\end{equation}
Here, $[]_{+}$ means keeping the positive part of the loss function and we want $S(f)>S(f')$.

\paragraph{Evaluation Metrics}
We use 4 metrics to evaluate our performance on the link prediction task. 
These are root mean square error (RMSE, $\sqrt{\frac{1}{n} \sum_{i=1}^{n}\left(\hat{y}_{i}-y_{i}\right)^{2}}$, where $\hat{y}_i$ is our predicted relation and $y_i$ is the true relation), Hits@K - the probability that our target value is in the top $K$ predictions, mean rank (MR) - the average ranking of each prediction, and mean reciprocal rank (MRR) to evaluate our performance on the link prediction task. These are standard metrics in the knowledge graph embedding community.
    
\section{Explainable Bias Detection}
The general problem is as follows. Given a knowledge graph $G = (V, \mathcal{F}, A)$, suppose that one attribute $a \in A$ is \emph{private}: we do not wish users of the embedded graph to be able to predict the value of $a$ for a given vertex $v \in V$. We show here that it is possible to predict the value of some private attributes even when that attribute is not used in the embedding algorithm.

We give three methods: a logistic classifier based method (detect-LC), Canonical Correlation Analysis (detect-CCA), and a linear decomposition method (detect-LD) to detect private information in the vertices $V$.

\begin{figure}[htbp]
\centering
\begin{minipage}{.45\textwidth}
    \includegraphics[scale=.21]{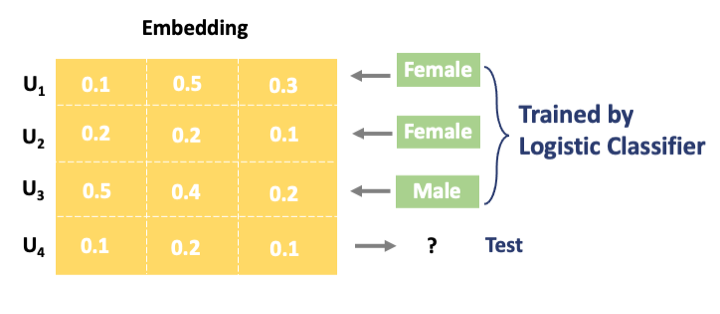}
    \caption{Schematic of detect-LC}
    \label{fig: Logistic Classifier}
\end{minipage}
\hfill
\begin{minipage}{.45\textwidth}
    \includegraphics[width=\textwidth]{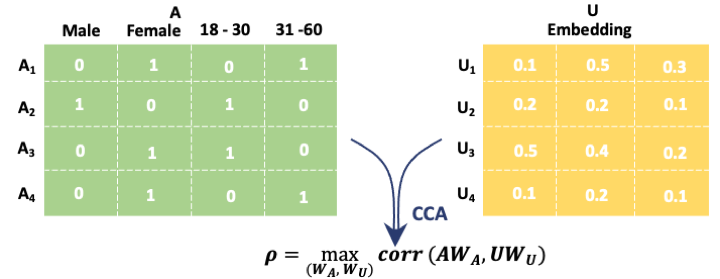}
    \caption{Schematic of detect-CCA}
    \label{fig: CCA}
\end{minipage}
\end{figure}

\paragraph{detect-LC: Logistic Classifier}
Consider a knowledge graph denoted by \( G = (V, \mathcal{F}, A) \). We've represented this graph in a \( d \)-dimensional space, \( \mathbb{R}^d \), based solely on the information from \( V \) and \( R \). Suppose  a subset of vertices $U \subseteq V$ is labelled with the presence or absence of attributes $a_i$. For example, in a database of users and movies, we might have a subset $U$ of users, with each node labelled as over 18 or under 18. We train a logistic classifier to predict the value of $a_i$ for each vertex $u \in U$. See Figure \ref{fig: Logistic Classifier} for an illustration. In the case of our movie rating example, private attributes could be gender, age, and occupation.

\begin{figure}[htbp]
\centering
\begin{minipage}{.60\textwidth}
    \includegraphics[scale=.19]{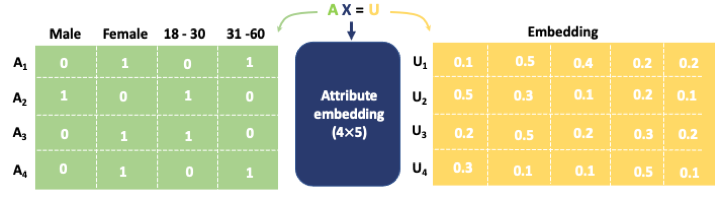}
    \caption{Schematic of detect-LD: Our Linear Decomposition System}
    \label{fig: linear}
\end{minipage}
\hspace{1.40cm}  
\begin{minipage}{.20\textwidth}
    \centering
    \includegraphics[scale=.20]{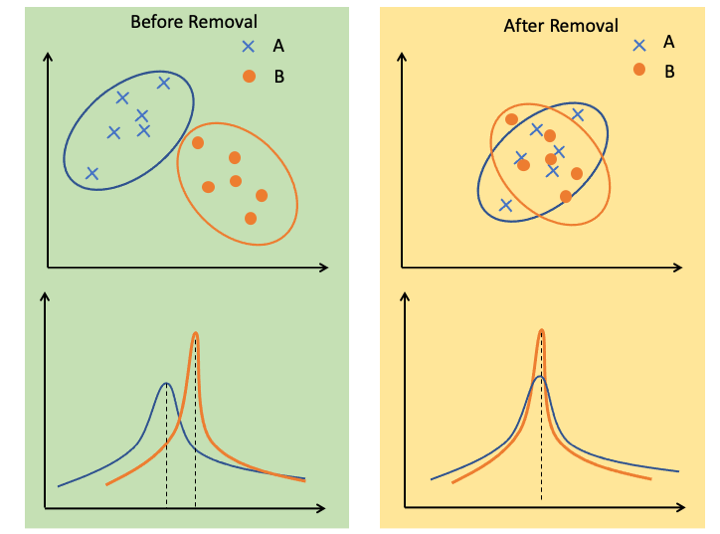}
    \caption{First Moment Loss}
    \label{fig: firstmoment}
\end{minipage}
\end{figure}

\paragraph{detect-CCA: Canonical Correlation Analysis}
Canonical Correlation Analysis (CCA) is used to measure the correlation information between two multivariate random variables \cite{shawe2004kernel}. Just like the univariate correlation coefficient, it is estimated on the basis of two aligned samples of observations.

As above, we assume that a subset $U\subseteq V$ of vertices is labelled with the presence or absence of attributes $a_i$. Supposing we have $n$ attributes, we can then form a Boolean-valued matrix $\mathbf{A}$ of dimension $|U|\times n$, where each row corresponds to a vertex in $U$, and each column corresponds to an attribute. In this matrix, each $u \in U$ is therefore represented by a vector of ones and zeros corresponding to presence or absence of each attribute. Suppose we have a  $|U|\times d$ matrix of vertex embeddings $\mathbf{U}$ that we have learnt via our knowledge graph embedding. We compute the CCA between $\mathbf{A}$ and $\mathbf{U}$. We learn vectors $\mathbf{w}_A$ and $\mathbf{w}_U$, such that the correlation between the projected $\mathbf{A}$ and projected $\mathbf{U}$ are maximised, that is 
$
    \rho =\max _{\left(\mathbf{w}_{A}, \mathbf{w}_{U}\right)} \operatorname{corr}\left( \mathbf{A} \mathbf{w}_A, \mathbf{U} \mathbf{w}_U\right)
    \label{eqn: correlation1} 
$.
Note there are $k$ correlations corresponding to $k$ components.

In the example of viewers and movies, we use this method to compare two descriptions of users, illustrated in Figure \ref{fig: CCA}. One matrix is based on demographic information, represented by Boolean vectors. The other matrix is based on their behaviour, computed by their movie ratings only.

\paragraph{detect-LD: Linear Decomposition}
\label{sec: linearsys}
Again assuming we have a matrix of entity embeddings $\mathbf{U}$ with Boolean matrix $\mathbf{A}$ encoding the attribute information, we investigate the possibility that the entity embeddings can be decomposed into a linear combination of embeddings corresponding to attributes. Specifically, we investigate whether we can learn a matrix $\mathbf{X}$ by solving $\mathbf{A}\mathbf{X} = \mathbf{U}$.

We investigate here the possibility that entity embeddings in knowledge graphs can be decomposed into linear combinations of embeddings corresponding to attributes. We use methods from 
\citet{xu2023on} to see if an entity embedding $\mathbf{u}$ can be decomposed.

In our example of viewers and movies, we define a set of users as $U$ and the Boolean encoding matrix of the attributes as $\mathbf{A}$. We aim to solve a linear system $\mathbf{A\mathbf{X}}=\mathbf{U}$, so that the user embedding can be decomposed into three components (gender, age, occupation) as follows, $\mathbf{u} = \sum_{i} a_i \mathbf{x}_i$, where $\mathbf{u}$ is a user embedding, $i$ ranges over all possible values of each private attribute, $\mathbf{x}_i $ is an embedding corresponding to the $i$th attribute value, and $a_i\in\{0, 1\}$ indicates the presence or absence of that attribute value. Figure \ref{fig: linear} shows a running example of decomposing user into gender and age (18-30 and 31-60). Figure \ref{fig: linear} shows the schematic of detect-LD.

\paragraph{Hypothesis Testing with Random Permutations}
\label{sec:stattest}
We test whether our methods are able to detect private information by comparing with randomly shuffled data. Our null hypothesis is that the embedding of a vertex $u$ and its attributes $a$ are independent. To test whether this is the case, we employ a non-parametric statistical test, whereby we directly estimate the $p$-value as the probability that we could obtain a ``good''\footnote{either high or low, depending on the statistic} value of the test statistic under the null hypothesis. If the probability of obtaining the observed value  of the test statistic is less that 1\%, we reject the null hypothesis. Specifically, we will
randomly shuffle the pairing of vertices and attributes 100 times, and compute the same test statistic. If the test statistic of the paired data is better than that of the randomly shuffled data across all 100 random permutations, we conclude that the correctly paired data performs better to a 1\% significance level. The test statistic for the logistic classifier is the accuracy of predicted labels, for CCA is the correlation $\rho$, and for the linear System $\mathbf{A}\mathbf{X}=\mathbf{U}$ we look at the L2 norm loss of the linear system, cosine similarity and retrieval accuracy, a metric defined in \cite{xu2023on}.

\section{Transparent Control of Bias}
\label{sec:removing}
As well as detecting bias in a knowledge graph, we also wish to remove it. In the following, we give two methods for removing bias based on a linear transformation method (remove-LP) as well as an extension of this (remove-LP-multi),
and two (remove-FM and remove-FM-multi) based on penalising differences between the distribution of items with different protected characteristics. The difference in distribution is approximated by comparing only the first moments, though the method can be extended to further moments.

\paragraph{(remove-LP): Linear Transformation} 
\label{sec:linearremoval}
 We refer to previous work on word embedding debias \cite{sutton2018biased} on eliminating sensitive attributes for user embedding. Suppose we have a knowledge graph $(V,\mathcal{F}, A)$ and an embedding of that graph, and we are interested in the attributes of a subset $U \subseteq V$ of the vertices. Suppose we have private attribute $a$ that we wish to protect, which takes two values $m$ and $n$. Let $U_m$ represnt a set of entity embeddings with attribute $m$, and $U_n$ is a set of entity embeddings with attribute $n$. If we identify a hyperplane that best separates these two sets of points, then $\hat{\mathbf{b}}$ can be defined as the normalised vector orthogonal to the hyperplane pointing from $m$ to $n$. Given a vertex embedding $\mathbf{u}$, we can project $u$ onto the hyperplane orthogonal to $\hat{\mathbf{b}}$ via $
\mathbf{u}_{\perp}={\mathbf{u}}-\left({\mathbf{u}} \cdot \hat{\mathbf{b}}^{T}\right) \cdot \hat{\mathbf{b}}
\label{eq: linear-debias}
$.

\paragraph{(remove-LP-multi): Linear Shifting for 2 or more classes}

For more than two classes, we propose a new linear-shifting method. Again suppose we have a graph $(V,\mathcal{F}, A)$, an embedding of this graph, and a subset $U\subseteq V$ vertices with particular attributes. Suppose that we are able to detect information about an attribute $a$ that we wish to keep private, and $a$ can take one of $n$ values. We take the set of entity embeddings $U = \{\mathbf{u}_i\}$, each having a value $a_j$ of the attribute $a$. For instance, in our running example of movie rating, $U$ may be users, $a$ may be occupation, and the values $a_j$ could be $\{$academic, artist, clerical, college student, ...$\}$. Now, the set $U$ of embeddings is partitioned into $n$ sets $\{U_1,..., U_n\}$ according to the value of $a$.
 We calculate the centroid $\mu_{j}$ of each of the sets $U_j$, and then calculate the centroid $\mu$ of all the centroids $\mu_j$. Finally, we shift each vector $\mathbf{u}_{jk} \in U_j$ by $\mu-\mu_j$, so that now, all the sets $U_j$ have the same centroid.

\paragraph{(remove-FM): First-Moment Loss}
\label{sec:firstmoment}
Ideally, two sets of items that differ only along protected dimensions should have the same distribution
in the embedding space. Therefore we aim to reduce the difference between distributions of items which
differ in that way by penalising the distance between first-moments of their empirical distribution (i.e. between their empirical means) during the training process.

Again suppose we have a graph $G = (V,\mathcal{F}, A)$, a subset of vertices of interest $U$, an embedding of this graph, and that we are able to detect information about an attribute $a$ that we wish to keep private. Consider two subsets of vertex embeddings $U_m = \{\mathbf{u}_{mi}\}$ and $U_n=\{\mathbf{u}_{nj}\}$ with private attribute $a$ taking values $m$ and $n$. For the first-moment loss we consider each of $U_m$ and $U_m$ as multivariate distributions of the two classes $m$ and $n$. We aim to remove the difference in the first moments of the two classes by centering the mean coordinates of the two classes. An illustration of this can be shown in Figure \ref{fig: firstmoment}. We follow \citet{zemel2013learning} and implement this method on our knowledge graph. In order to do this we minimize the following loss function:
\begin{equation}
  \textbf{L}  = \text { Loss }
     +\sigma\left\|\frac{g_m}{N_{m}}\sum_{i} \mathbf{u}_{mi} +\frac{g_n}{N_{n}}\sum_{j} \mathbf{u}_{nj}\right\|
\end{equation}
where $g_{i} \in \{-1,1\}$, whilst $g_m = 1, g_n = -1$. $N_{m}, N_{n}$ stand for the counts of vertices with attribute taking values $m$ and $n$.  $\sigma$ is a weight parameter that gives more weight to the regularization, Loss refers to Equations \ref{eq:actual_loss}, \ref{eq: actual_loss_margin} for predicting relations/entities. Consider the example of knowledge graph of viewers and movies, we aim to neutralize gender information in the first moment of the two multivariate distributions by standardizing the mean coordinates across genders.

\paragraph{(remove-FM-multi): First-Moment loss for 2 or more classes}
For more than two classes, we adapt the ``mean-match" \cite{kamishima2012enhancement} process. We impose a constraint, limiting the norm of the embeddings to 1. Our rationale for this is rooted in the belief that this constraint might preserve or even boost the predictive capability of the model. We aim to test this conjecture and, if corroborated, demonstrate its implications for the robustness of the predictive model. We propose a new loss function in \ref{eq: remove-FM-multi},
\begin{equation}
\begin{aligned}
&\begin{aligned}
\textbf{L} = \theta*\text { Loss } & +\sum_{i}\frac{1}{N_{c_{i}}}\left(\|\mathbf{u}_{c_{i}}-\mu_{c_i} \|^2-1\right)^2 
& +\sum_{i j}\frac{1}{N_{c_{i,j}}}\left(\mu_{c_i}-\mu_{c j}\right)^2
\end{aligned}\\
\end{aligned}
\label{eq: remove-FM-multi}
\end{equation}
where $c_{i, j} \in\{\text {Class 1, Class2, Class3 ... }\}$, $N_{c_{i}}$ is the count of vertices of class $i$, $N_{c_{i,j}}$ is the count of combinations of different classes. $\mu_{c_i}$ is the center point of class $i$, $\theta$ is the weight parameter that gives more weight to the Loss (the original loss in Equation \ref{eq:actual_loss}, \ref{eq: actual_loss_margin}) for predicting relations/entities.

\section{Experimental Study}
\subsection{Datasets and Training Details}
\paragraph{MovieLens}
\label{sec:movielenstraining}
This experiment was conducted on the MovieLens 1M dataset \cite{harper2015movielens} which consists of a large set of movies and users, and a set of movie ratings for each individual user. It is widely used to create and test recommender systems.

Users and movies each have attributes. For example, users have demographic information such as gender, age, or occupation. Whilst this information is typically used to improve the accuracy of recommendations, we use it to test whether the embedding of a user correlates to private attributes, such as gender or age. Crucially, \textbf{we compute our graph embedding based only on ratings, leaving out user attributes}. Experiments for training knowledge graph embeddings are implemented with the OpenKE \cite{han-etal-2018-openke} toolkit. 

We use a 90:10 train:test split to train user and movie embeddings with triples $(user, rating, movie)$. Initial embeddings are randomly assigned and optimized to minimize loss as per equation \eqref{eq:actual_loss}. We sample 10 corrupted entities and 4 corrupted relations for each true relation. The learning rate is 0.01 and epochs are set at 300. Link prediction on the 10\% test set yields an RMSE score of 0.88 (table \ref{tab: performance comparision}).

\paragraph{Citation Network}
The KG20C knowledge graph is constructed from 20 top AI conferences \cite{tran2019exploring}. The entities consist of 5407 papers, 8060 authors, 692 affiliations, 1920 domains, and 20 conferences. We train entity embeddings for papers, authors, affiliations, and domains, and relation embeddings of the following four types: ``Author in affiliation", ``Author write paper", ``Paper cite paper" and ``Paper in domain".

We view the venue, i.e. conference, as information that we wish to detect and potentially remove. As an application, suppose we are building a search engine and wish to avoid the ``echo chamber'' of groups of authors citing each other's work. Detecting conference information could help to return a more diverse set of search results.

Embeddings for entities and relation types were randomly initialized and trained to minimize the loss as per equation \ref{eq: actual_loss_margin}, using a 0.05 learning rate over 300 epochs and a margin, $\gamma$, set to 6. Link prediction task metrics (Hits@10: 0.29, Hits@1: 0.075, MRR: 0.15, and MR: 1369.93) attest to our embeddings' quality. The bias-removal strength, $\sigma$, incrementally set to 50, 100, and 150, increased the loss function's bias-removal component. With 1/3 of triples sharing the same head and relation but different tails, individual triples were tested to maintain accuracy. To ensure the reproducibility of our results, we have made the code available at \url{https://github.com/ZhijinGuo/EXTRACT}.

\subsection{MovieLens Bias Detection}
We now apply our three methods for bias detection to investigate the extent to which private information can be detected in user embeddings trained without that information. 
\paragraph{detect-LC on MovieLens}
\label{sec:logclass}
We use a logistic classifier to predict gender, age, and occupation from the embeddings. We sub-sample the data to have balanced datasets with respect to gender, age, and occupation.

Using an 80:20 training-to-test split across all datasets, we trained logistic classifiers for each attribute, implemented via the scikit-learn toolkit \cite{scikit-learn}. Preliminary results highlight the model's ability to predict private traits, with 73\% accuracy in predicting gender, approximately 72\% and 68\% in predicting younger and older users respectively, and 59\% accuracy for occupation prediction. These binary classification results outperform the 0.50 baseline.

\paragraph{detect-CCA on MovieLens}
We collect attribute information for all 6040 users and represent their personal attributes with Boolean indicator vectors $\mathbf{a}_i$ which encode the value of each attribute (gender, age, and occupation). See figure \ref{fig: CCA} for a schematic of these vectors.

\begin{figure}[htbp]
    \centering
    \begin{subfigure}{.45\textwidth}
        \centering
        \includegraphics[scale=.40]{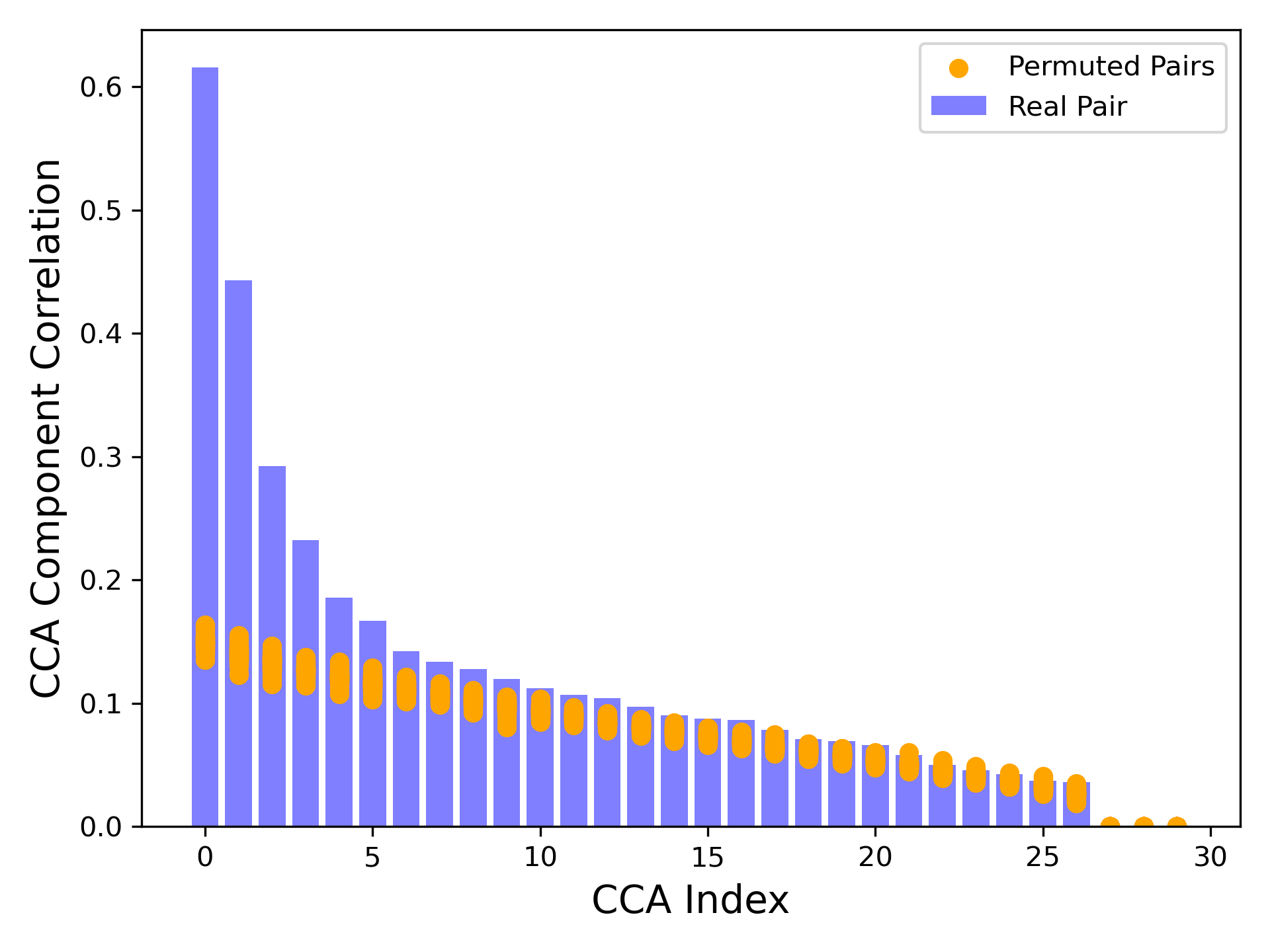}
        \caption{Pearson's correlation coefficient (PCC) for true user-attribute pairings and 100 permuted pairings. PCC is calculated between projected $\mathbf{A}$ and projected $\mathbf{U}$. $x$ axes stands for the $k$th components, $y$ axes gives the value. The PCC value for real pairings is larger than for any permuted pairings.}
        \label{fig: p-value-kg}
    \end{subfigure}
    \begin{subfigure}{.45\textwidth}
        \centering
        \begin{adjustbox}{valign=b}
            \includegraphics[scale=.25]{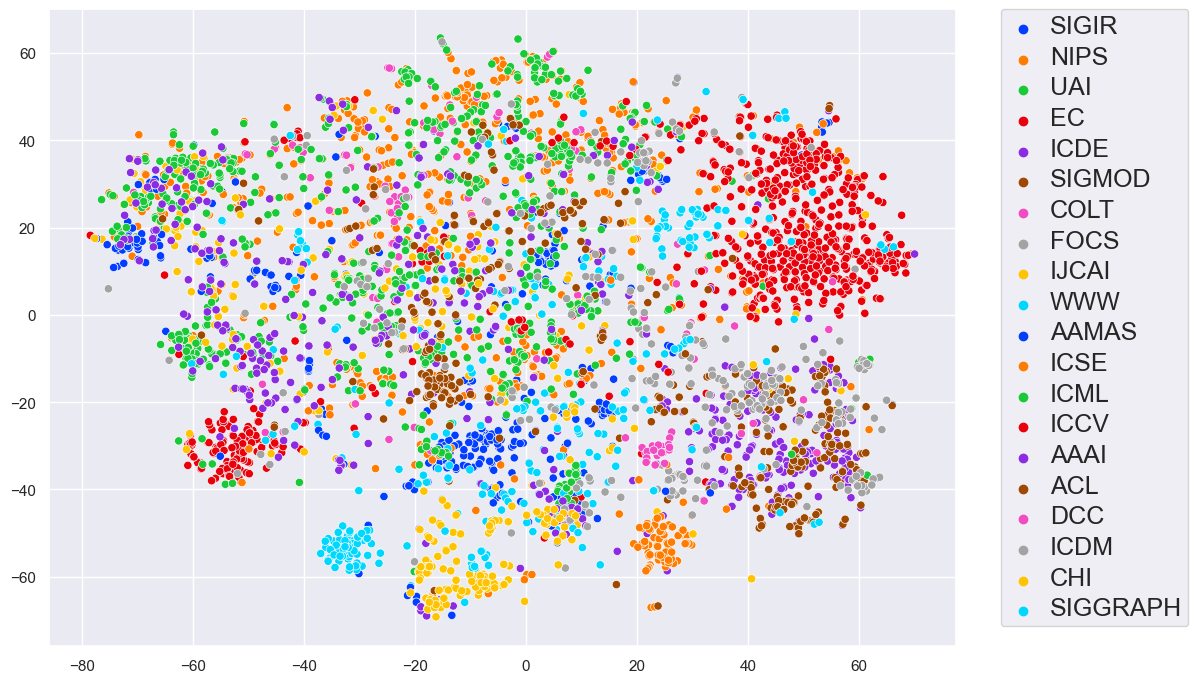}
        \end{adjustbox}
        \caption{(t-SNE) visualisation \cite{van2008visualizing} for paper embedding, colour-coded by conference. Notice that even though we do not use conference information in learning these embeddings, information about the conferences is clearly present in the embeddings.}
        \label{fig: tsne-conference}
    \end{subfigure}
    \caption{Explainable Detection of Bias}
    \label{fig:subfigures}
\end{figure}

We apply CCA to calculate the correlation between users and their attributes. We apply the non-parametric statistical test described in section \ref{sec:stattest}. Specifically, our null hypothesis is that users' movie preferences are not correlated with their attributes. We calculate Pearson's correlation coefficient (PCC) between projected $\mathbf{A} \mathbf{w}_A$ and projected $\mathbf{U} \mathbf{w}_U$. We go on to calculate the PCC between 100 randomly generated pairings of user and attribute embeddings, and find that the PCC between true pairs of attribute and user embeddings is higher each time. We therefore reject the null hypothesis at a 1\% significance level. 
The correlation coefficients between real pairs and random pairs is reported in figure \ref{fig: p-value-kg}. 

\paragraph{detect-LD on MovieLens}
We investigate the ability of a user embedding to be reconstructed as a linear sum of attribute embeddings. We find that a user embedding can be reconstructed as a linear combination of its attributes by solving the linear system described in section \ref{sec: linearsys} with pseudo-inverse method.
In order to interpret the user embedding with user attributes such as gender, age and occupation, we first group the user by age and gender firstly and compute the mean embedding of 14 group of users. We afterwards group the user by gender, age and occupation and compute the mean embedding of 241 group of users. Two test statistics are used to test our linear system with significance threshold $\alpha = 0.01$.

As with the CCA setting, we permuted the pairing of users 100 times.
Table \ref{tab: statistic test} shows the observed p-value for three different statistics, which is the probability of seeing that value of statistic under the null hypothesis.
We first decompose the user embedding into gender and age. Our results show the linear system is able to decompose the user embedding with a loss of 0.47, which is lower than every loss for a random permutation (1.11-2.11). The cosine similarity is 99.8\%, higher than any permuted pairs. The identity retrieval accuracy is 0.79 which is higher than any random permuted pairs (0.0-0.14). Therefore, the null hypothesis is rejected. This shows that a user embedding can be reconstructed as a linear combination of gender and age.

When decomposing the embedding into gender, age and occupation, the L2 norm is 17.87 which is lower than every loss for a random permutation (18.90-19.56). The cosine similarity is 97.1\%, higher than any permuted pairs. As for identity retrieval accuracy, although the value is only 0.23 which is not a good result, it is still higher than any random permuted pairs (0.00-0.08). Therefore, the null hypothesis is rejected.

\begin{table*}
\centering
    \caption{p-value for hypothesis test. Note that * indicates better than random baseline to significance level $\alpha$ = 0.01. In our case, we are estimating directly the p-value, as the probability of an event, that we could have a high (low) value of the test-statistic by chance under the null-hypothesis}
    \begin{tabular}{@{}llllll@{}}
    \hline
                 & L2 Norm & Cosine Similarity & Retrieval Acc. & p-value \\ \hline
    Gender, Age Real Pair   & 0.47* & 99.8\%   & 0.79* & <0.01  \\ \hline
    Gender, Age Permuted    & 1.11-2.11*  & 96.5\%-99.0\%  & 0.00-0.14* &  <0.01  \\ \hline
    Gender, Age, Occ Real Pair& 17.87* &  97.1\%   &    0.23* &  <0.01 \\ \hline
    Gender, Age, Occ Permuted & 18.90-19.56* & 96.2\%-96.8\%  &    0.00-0.08* &  <0.01 \\ \hline
    \end{tabular}
    \label{tab: statistic test}
\end{table*}

\subsection{Citation Network Bias Detection}
We consider a dataset of scientific papers, whose entities are the papers, their authors, the conference in which those papers were presented, and the scientific domain they belong to. For the sake of example, we will treat as sensitive the information about the specific conferences in which a given paper was presented (we could pretend that this information should not be used
as part of career assessment, and therefore should be hidden for this reason).

We embed scientific papers using their relations with the sets of authors and domains, and then we will test if that embedding could contain enough information for a logistic classifier to learn how to predict the specific conference. Note that this is a very strict requirement - that in no way can any information relevant to this task be contained in the embedding of the paper. Similar to the movie recommender system, we treat the conference information in this citation network as a sensitive attribute.

Utilizing 5407 papers with respective author, citation, domain, and author affiliation relations, we generate embeddings for each and label them with one of the 20 conferences, which are then split into training and test sets. Our logistic classifier (detect-LC) yields a 55\% accuracy in conference venue prediction, surpassing the 13\% baseline (\ref{fig: tsne-conference}). Despite not being trained on conference data, the embeddings can still distinguish different conferences.

\subsection{Transparent Control of Bias}
We apply the techniques introduced in section \ref{sec:removing} to remove information about age and gender from MovieLens-1M, and information about conference from the KG20C citation network.

\paragraph{remove-LP for MovieLens}
\label{sec: movielens linear}
\begin{figure}
	\centering
	\begin{subfigure}{.35\columnwidth}
		\centering
		\includegraphics[width=\textwidth]{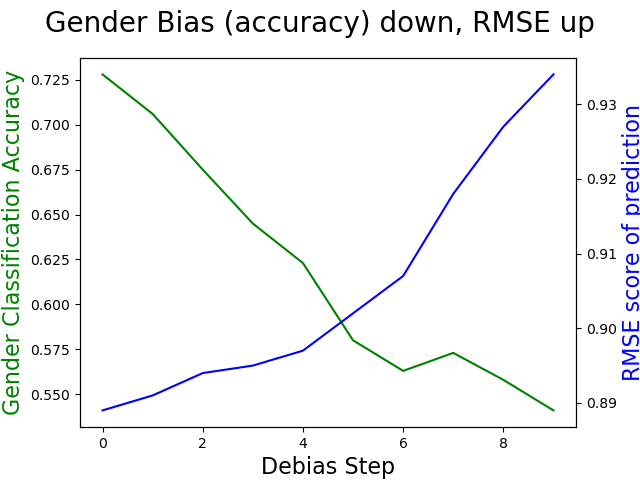}
         \caption{Gender - rating trade-off}
         \label{fig: gender-tradeoff}
	\end{subfigure}
 \hspace{1cm}
	\begin{subfigure}{.35\columnwidth}
		\centering
		\includegraphics[width=\textwidth]{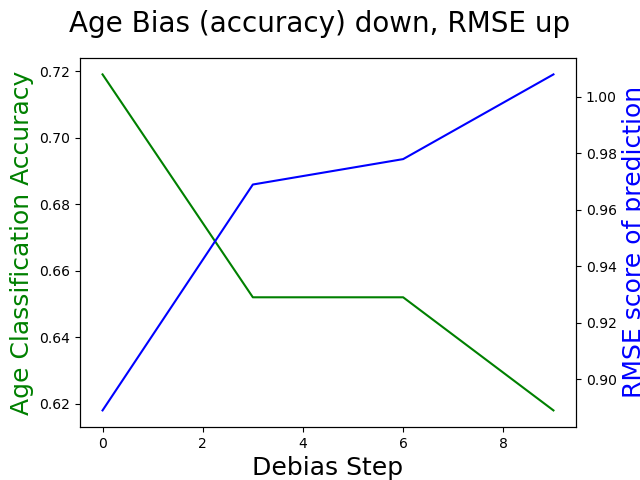}
         \caption{Age - rating trade-off}
          \label{fig: age-tradeoff}
	\end{subfigure}
 \caption{Bias - rating prediction trade-off, model becomes gender neutral with slightly losing prediction power}
\end{figure}

We investigate the ability to remove the gender and age bias detected in the knowledge graph user embedding whilst maintaining the ability of the embedding to predict behaviour. We apply remove-LP (section \ref{sec:linearremoval}) to remove gender and  age information from the data splits described in section \ref{sec:logclass}. We run the bias removal process for 10 iterations. Results are reported in table \ref{tab: performance comparision}.

As depicted in figures \ref{fig: gender-tradeoff} and \ref{fig: age-tradeoff}, there is a trade-off between predictive power and bias. An increase in iterations leads to a drop in gender classification accuracy from 73\% to 54\%, nearing baseline accuracy, with a slight increase in RMSE from 0.88 to 0.93. A similar pattern is observed in age bias removal, with accuracy for under/over 50s decreasing from 0.72 to 0.62, and RMSE rising from 0.88 to 1.01.

\paragraph{remove-FM for MovieLens}
\label{sec: movielens moment}
We use remove-FM  (section \ref{sec:firstmoment}) to remove bias during the training of the embeddings. We again use embeddings and splits described in section \ref{sec:logclass}, and investigate how well predictive power is retained as bias information is removed. Figures are reported in table  \ref{tab: performance comparision}. We see that as we increase the parameter $\sigma$, the accuracy of the gender classifier decreases, although not monotonically. The RMSE increases slightly from 0.88 to 0.96.

\paragraph{remove-LP-multi for KG20C}
Regarding the KG20C dataset, we employ linear shifting for classes greater than or equal to two to eliminate bias post-training, adjusting the paper embedding to ensure that paper sets labelled with different conferences converge at a common center. Results, detailed in Table \ref{tab: performance comparision KG20C}, indicate that the model, while becoming conference neutral, experiences a minor decrease in predictive power (Hits@10 reduces from 0.29 to 0.21).

\paragraph{remove-FM for KG20C}
We apply the methods described in the multi-class version of remove-FM (section \ref{sec:firstmoment}). The embeddings are debiased during training.
A parameter $\theta$ weights the strength of the prediction power. We set $\theta$ to 1 and 20, to increase the influence of the prediction part of the loss function. 
As shown in Table \ref{tab: performance comparision KG20C}, a larger $\theta$ slightly improves Hits@10 but decreases Hits@1 and MR. Additionally, the retained conference information increases accuracy from 0.14 to 0.26, though Hits@10 drops 0.14, signifying a trade-off between prediction power and unwanted information removal.

\begin{table}[h]

\begin{center}
\caption{Performance Comparison for Gender and Age Bias Removal, 
Mean accuracies for the gender and age logistic classifiers are reflected in the respective gender and age rows. Random refers to untrained random embedding, KG refers to the non-debiased embedding, while 50, 100, 150 represent for the value of $\sigma$, \underline{underline} numbers denote mean accuracy of logistic classifier for embedding without debiasing. We compare our results with the best results from the literature, ICML\_2019  \cite{bose2019compositional}, and FairGo \cite{wu2021learning}.}
\begin{tabular}{l|c|c|c|c|c|c|c|c|c}
\hline
 & Random & KG & \multicolumn{2}{|c|}{remove-LP}  & \multicolumn{3}{|c|}{remove-FM50/100/150} & ICML\_2019 & FairGo\\ 
\hline
RMSE & 1.26 & 0.88 & 0.93 & 1.01 & 0.96 & 0.96 & 0.96 & 0.92 & 0.92\\
Gender & 0.50 & \underline{0.73} & 0.54 & - & 0.66 & 0.62 & 0.58 &0.53 &0.50 \\
Age & 0.50 & \underline{0.72} & - & 0.62 & - & - & - & - & -\\ \hline
\end{tabular}
\label{tab: performance comparision}
\end{center}
\end{table}

\begin{table}[h]
\begin{center}
\caption{Performance Comparison for Conference Bias Removal, The mean accuracies for conference logistic classifiers are reflected in the respective conference column. Random refers to untrained random embedding, KG refers to the embedding without debiasing, \underline{underline} numbers denote mean accuracy of logistic classifier for embedding without debiasing. 1, 20 represent for the value of $\theta$.}

\begin{tabular}{@{}lllllllcc@{}}
\hline
  & Hits@1 & Hits@10 & MRR & MR & Conference  \\ 
 \hline
Random & 0.00  & 0.00 & 0.00 & 4106.62 &0.15 \\
KG & 0.075  & 0.29 & 0.15 & 1369.93 & \underline{0.55}  \\
remove-LP-multi &0.053  & 0.21 & 0.11 & 2044.66 & 0.10\\
remove-FM-multi1 & 0.05  & 0.15 & 0.09 & 971.57 & 0.14\\
remove-FM-multi20 & 0.04  & 0.17 & 0.09 & 1466.61 & 0.26\\ 
\hline
\end{tabular}
\label{tab: performance comparision KG20C}
\end{center}
\end{table}

\section{Discussion and Conclusions}
In this work, we introduced EXTRACT, a comprehensive framework that combined three methods for detecting and four methods for removing biases in knowledge graph embeddings. Using datasets like MovieLens 1M and KG20C, EXTRACT efficiently identified leaks of sensitive attributes and unintended conference affiliations. This two-step process ensured that specific dimensions of user embeddings, correlating to private or sensitive attributes, were identified and appropriately treated. Our findings showed that the correlations between the private attributes and the user representations were significantly higher than random, validating the efficacy of EXTRACT in capturing statistical patterns that reflect private attribute information.

Furthermore, our linear decomposition system, an integral part of the EXTRACT framework, revealed that user-behaviour-embedding can be approximated by a sum of user-demographic vectors. This validated the framework's ability to decompose user embeddings into a weighted sum of attribute embeddings, thereby allowing for the targeted removal or control of bias.

Our four approaches to debiasing were more transparent and explainable than more complex methods. Remove-LP method was most successful at debiasing, whilst retaining relatively good performance than the link prediction task. Overall, EXTRACT served as a unified, two-step approach that advanced the field towards ethical and transparent knowledge graph embeddings. 

\newpage

\bibliography{main}

\end{document}